\documentclass[letterpaper, 10 pt, conference]{ieeeconf}  

\usepackage{tabularx}   
\usepackage{siunitx}    
\usepackage{booktabs}   
\usepackage{array}
\usepackage{multirow}
\usepackage{graphicx}
\usepackage{comment}
\usepackage{amsmath,amssymb}
\usepackage{color}
\usepackage{url}
\usepackage{hyperref}
\newcommand{\ourmethod}{MonoCT\xspace}%
\newcommand*{\inparagraph}[1]{\noindent\textbf{#1}\hspace{0.5em}}

\usepackage{multirow} 
\usepackage{makecell} 

\usepackage{colortbl}
\usepackage{tabularx}
\usepackage{amsmath}
\usepackage{svg}
\usepackage{amsmath}
\usepackage{graphicx}
\usepackage{subcaption}
\usepackage{wrapfig}

\usepackage[capitalize]{cleveref}

\RequirePackage{xspace}
\makeatletter
\DeclareRobustCommand\onedot{\futurelet\@let@token\@onedot}
\def\@onedot{\ifx\@let@token.\else.\null\fi\xspace}

\usepackage{etoolbox,siunitx}
\robustify\bfseries
\usepackage{amssymb}
\usepackage{pifont}
\usepackage[nohyperlinks, printonlyused, withpage, smaller]{acronym}
\acrodef{KDE}[KDE]{\emph{kernel density estimator}}
\acrodef{iou}[IoU]{\emph{intersection over union}}
\acrodef{mle}[MLE]{\emph{maximum likelihood estimation}}
\acrodef{roi}[RoI]{\emph{region of interest}}
\acrodef{fov}[FOV]{\emph{field of view}}
\acrodef{bev}[BEV]{\emph{bird's-eye view}}
\acrodef{wbf}[WBF]{\emph{weighted box fusion}}
\acrodef{pls}[PLS]{\emph{Pseudo Label Scoring}}
\acrodef{em}[EM]{\emph{Ensemble Merging}}
\acrodef{dm}[DM]{\emph{Diversity Maximization}}
\acrodef{gde}[GDE]{\emph{Generalized Depth Enhancement}}
\acrodef{cvis}[CVIS]{\emph{Cooperative Vehicle-Infrastructure Systems}}

\usepackage{footnote}
\usepackage{fancyhdr}
\usepackage[absolute]{textpos}
\def\eg{\emph{e.g}\onedot} 
\def\Eg{\emph{E.g}\onedot}

\def\cf{\emph{cf}\onedot}

\def\wrt{w.r.t\onedot}

\def\etal{\emph{et al}\onedot}
\makeatother

\IEEEoverridecommandlockouts                              

\title{\LARGE \bf
\ourmethod: Overcoming Monocular 3D Detection Domain Shift with Consistent Teacher Models
}

\author{Johannes Meier$^{^*,1,2,3}$, Louis Inchingolo$^{^*,2}$, Oussema Dhaouadi$^{1,2,3}$,  \\ Yan Xia$^{2,3,\dagger}$, Jacques Kaiser$^{1}$, Daniel Cremers$^{2,3}$ \\
$^{1}$ DeepScenario\quad $^{2}$ TU Munich\quad $^{3}$ Munich Center for Machine Learning
\thanks{*Equal Contribution, $\dagger$ Corresponding author.}%
\thanks{TUM: {\tt j.meier@tum.de}, {\tt firstname.lastname@tum.de}}%
\thanks{DeepScenario: \tt firstname@deepscenario.com}
\thanks{This work is a result of the joint research project STADT:up. The project is supported by the German Federal Ministry for Economic Affairs and Climate Action (BMWK), based on a decision of the German Bundestag. The author is solely responsible for the content of this publication. This work was also supported by the ERC Advanced Grant SIMULACRON. We acknowledge ChatGPT for assistance with polishing and grammar checks.}%
}

\begin{document}
\begin{textblock}{175}(16,263)   
\scriptsize
\framebox{\parbox{\textwidth}{
Accepted at IEEE International Conference on Robotics and Automation (ICRA) 2025. \copyright 2025 IEEE.  Personal use of this material is permitted.  Permission from IEEE must be obtained for all other uses, in any current or future media, including reprinting/republishing this material for advertising or promotional purposes, creating new collective works, for resale or redistribution to servers or lists, or reuse of any copyrighted component of this work in other works.”
}}
\end{textblock}

\maketitle
\thispagestyle{empty}
\pagestyle{empty}

\begin{abstract}

We tackle the problem of monocular 3D object detection across different sensors, environments, and camera setups. In this paper, we introduce a novel unsupervised domain adaptation approach, \ourmethod, that generates highly accurate pseudo labels for self-supervision. Inspired by our observation that accurate depth estimation is critical to mitigating domain shifts, \ourmethod introduces a novel Generalized Depth Enhancement (GDE) module with an ensemble concept to improve depth estimation accuracy. 
Moreover, we introduce a novel Pseudo Label Scoring (PLS) module by exploring inner-model consistency measurement and a Diversity Maximization (DM) strategy to further generate high-quality pseudo labels for self-training. Extensive experiments on six benchmarks show that \ourmethod outperforms existing SOTA domain adaptation methods by large margins ($\sim$\textit{21\% minimum} for AP Mod.) and generalizes well to car, traffic camera and drone views. 

\end{abstract}

\section{INTRODUCTION}
Autonomous driving systems depend on robust perception to navigate safely, making accurate 3D object detection essential for identifying traffic participants (e.g., vehicles)~\cite{xia2021vpc}. Although LiDAR-based detectors \cite{shan2023scp,largekernel3d,focalformer3d, xia2023lightweight} provide high precision due to the accurate depth measurement, they are costly and require complex infrastructure. 3D object detection from monocular cameras offers a more affordable and accessible alternative, drawing increasing attention in the robotics and computer vision communities~\cite{wysocki2023scan2lod3}.

\begin{figure}[t]
\includegraphics[width=1.0\linewidth]{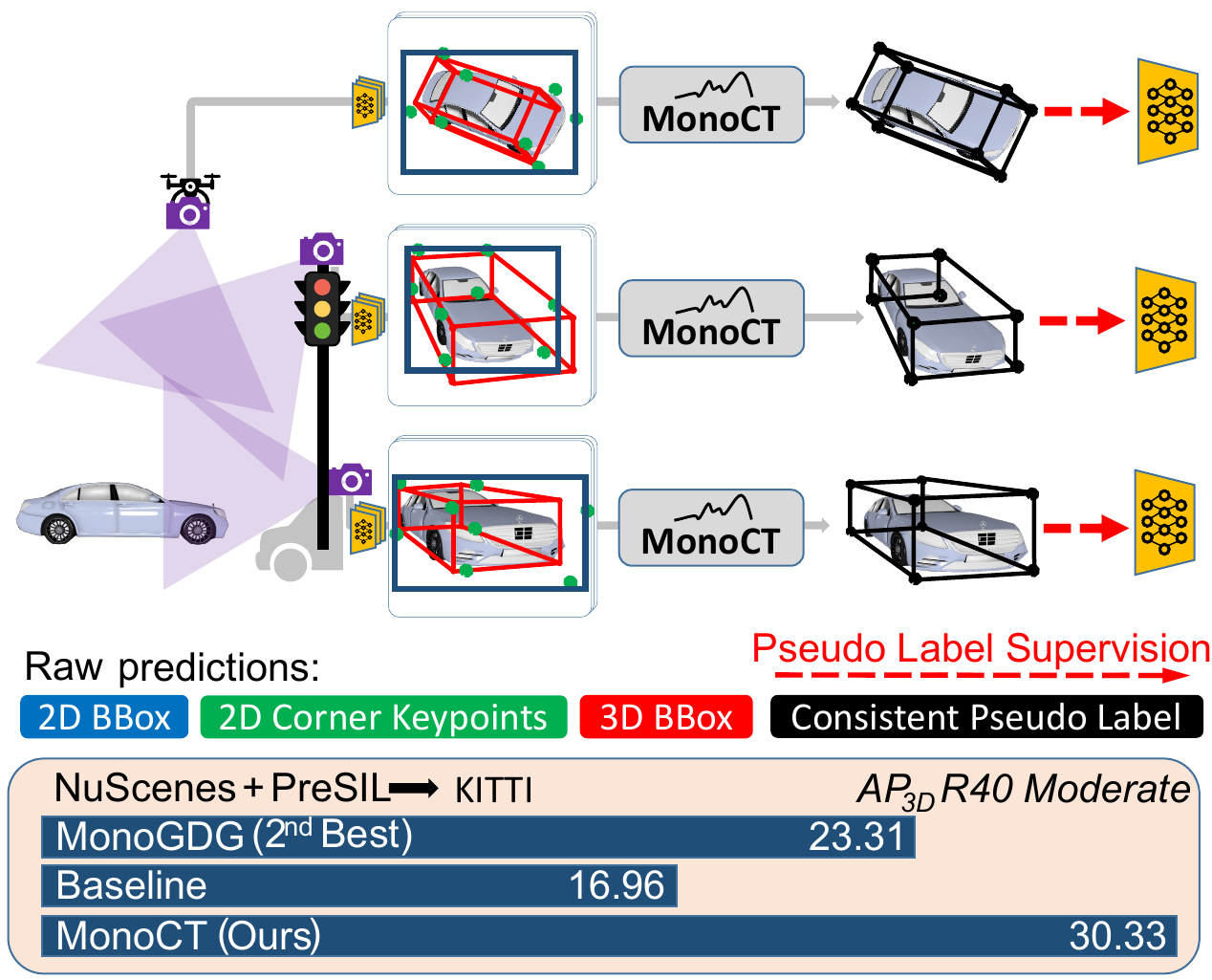}
  \caption{\ourmethod generates precise pseudo labels to overcome domain shift in monocular 3D object detection, utilizing inner-model and multi-model consistency. These labels are then used to self-supervise the detector. Our approach accommodates car-traffic-drone views and greatly improves the current SOTA methods.}
  \label{fig:teaser}
  \vspace{-0.7cm}
\end{figure}

Recent monocular 3D detectors work on the RGB images captured from the cameras mounted on autonomous vehicles. With the development of \ac{cvis}, traffic cameras and drone data are becoming increasingly important \cite{v2x-i,rope3d,cdrone}. Traffic cameras, fixed in position, can improve scene understanding \cite{v2x-i, v2x-seq}, while drones can capture occlusion-free trajectory data, that can be used to enhance planning and prediction models for autonomous driving \cite{highD,inD}. 
It is therefore essential to develop the capability to perform monocular 3D detection across car-traffic-drone views, as shown in \cref{fig:teaser} (Top).

We find that current monocular 3D detection methods suffer performance drops when used with different sensors, weather conditions, geographic locations, annotation styles, or traffic densities.
For instance, CoBEV \cite{cobev} and BEVFormer \cite{bevformer}) exhibit performance drops of 57.4\% and 48.7\%, respectively, when applied in unknown versus known camera locations (\cf Rope3D Homologous vs. Rope3D Heterologous) \cite{rope3d,cobev}. While domain adaptation has been widely studied for LiDAR data \cite{see-vcn,uncertainty_aware_mean_teacher, exploiting_playbacks,ijcai2024p72}, to date, a few monocular 3D detection methods have been proposed for domain shift in different environments and camera views. 
MonoGDG~\cite{monogdg} and DGMono3D~\cite{dgmono3d} focus on sensor alignment, calibrating the \ac{fov} between source and target domains.
STMono3D~\cite{stmono3d} and MonoTTA~\cite{monotta} explore a self-training strategy and simple filter schemes for pseudo-label selection. However, they overlook the importance of depth and fail to identify high-quality pseudo labels.
Recent Mix-Teaching \cite{mix-teaching} explores a semi-supervised way to train an ensemble of teacher models. However, it selects pseudo labels without considering the inner-model depth consistency. 
To overcome these limitations, we propose a novel network, named \ourmethod, to address the Monocular 3D detection domain shift across car-traffic-drone views with Consistent Teacher (CT) models. 

As pointed out in \cite{monoflex, monocd, monodde}, the main challenge in monocular 3D object detection lies in accurately predicting depth. Inspired by that, 
we first propose a novel ensemble-based method to improve the quality of depth estimation. Specifically, we generate 48 auxiliary depth estimates using predictions of 2D keypoints and 2D bounding boxes. These are then consolidated into a single, outlier-resistant prediction using kernel density estimation. We then propose a Pseudo Label Scoring (PLS) module to evaluate the quality of pseudo-labels by leveraging both 2D/3D consistency and inner-model depth consistency. Another observation is that cars with orientations driving forward, backward, left, or right relative to the camera are selected more frequently as pseudo labels.
To avoid oversampling and bias toward simpler cases, we enhance the generation of highly accurate and diverse pseudo-labels by introducing Diversity Maximization.

To summarize, the key contributions of this work are:

\begin{itemize}
    \item We focus on the fairly understudied problem of monocular 3D object detection across car-traffic-drone views, to overcome the domain shift challenges.
    \item We propose a novel \ac{gde} method that can accurately estimate depth and generalize effectively to car-traffic-drone perspectives.
    \item We propose a \ac{pls} module and show that it can predict the overlap of the top pseudo labels with ground truth more accurately than previous scoring methods.
    \item We design a \ac{dm} strategy to avoid model bias towards simpler configurations when selecting pseudo labels.
    \item We conduct extensive experiments on the six widely used benchmarks, NuScenes, Lyft, KITTI, PreSIL, Rope3D, and CDrone, including cross-dataset, synthetic-to-real, and weather condition domain shifts. Experiments demonstrate that the proposed \ourmethod outperforms the state-of-the-art methods.
    
\end{itemize}


\section{Related Works}
\inparagraph{Monocular 3D Object Detection}
Recent monocular 3D detection research has centered on egocentric car-view datasets like KITTI \cite{kitti} and Waymo \cite{waymo}, with depth estimation being a key challenge \cite{monoflex, monoground, monocd}. 
While LiDAR point clouds improve depth accuracy \cite{monopgc,dd3dv2,xia2021vpc}, this dependency limits applicability to datasets without LiDAR \cite{rope3d,v2x-i}.
DEVIANT \cite{deviant} and MonoLSS \cite{monolss} explore depth-equivariant backbones and feature relevance scoring to enhance depth without LiDAR.
Yan et al. \cite{monocd} showed that ensemble-based depth estimation improves detection accuracy. 
Several works use 2D keypoints to aid depth estimation: MonoFlex \cite{monoflex} and MonoDDE \cite{monodde} derive depth from keypoints, while RTM3D \cite{rtm3d} and MonoCon \cite{monocon} leverage keypoints for object positioning and generalization. 
However, these methods are limited to car views due to assumptions like a single y-axis rotation \cite{cdrone}. In contrast, our \ourmethod generalizes across drones and traffic views.

\inparagraph{Domain adaptation for LiDAR based 3D object detection}
Recent research in domain adaptation for 3D object detection has primarily focused on LiDAR-based detectors. 
Current methods address domain shift via sensor alignment \cite{see-vcn,uav3d}, feature alignment \cite{uav3d,stal3d}, student-teacher training \cite{uncertainty_aware_mean_teacher,li2024vxp,mlc-net,pere,stal3d}, and temporal consistency \cite{ms3d++,soap,exploiting_playbacks}. 
MS3D \cite{ms3d} and MS3D++ \cite{ms3d++} use kernel density estimation (KDE) to combine predictions from multiple teacher models trained on distinct source domains.
While we also employ a KDE, we apply it to depth estimation. Image-based detectors like BEVUDA \cite{bevuda} and MonoTDP \cite{monotdp} incorporate LiDAR during training to handle domain shifts, but we avoid this dependency since not all monocular datasets include LiDAR \cite{rope3d}.

\inparagraph{Domain adaptation for monocular 3D object detection}
Current cross-dataset adaptation methods normalize camera intrinsics to handle varying parameters \cite{cross_dataset_sensor,dgmono3d,stmono3d,monogdg}.
Zheng \etal \cite{cross_dataset_sensor} align the ego-frame and use extrinsics-aware RoI sizes, while DGMono3D \cite{dgmono3d} and MonoGDG \cite{monogdg} align the \ac{fov} between source and target domains, with DGMono3D also utilizing target dataset object dimensions. 
Similar to STMono3D \cite{stmono3d}, we leverage unlabeled target data to enhance accuracy, though we show standard 3D detection scores fail to identify high-quality pseudo labels. 
WARM3D \cite{warm-3d} uses weak supervision with 2D bounding box labels and a 2D/3D consistency loss, focusing on traffic views. 
In this work, we solve domain shifts in all perspectives, including car, traffic camera and drone views.
While Mix-Teaching \cite{mix-teaching} filters based on teacher ensemble agreement, we show that filtering based on inner-model consistency is more effective.

\section{Method}

\begin{figure*}

\begin{minipage}[c]{0.70\linewidth}
\includegraphics[width=1.00\linewidth]{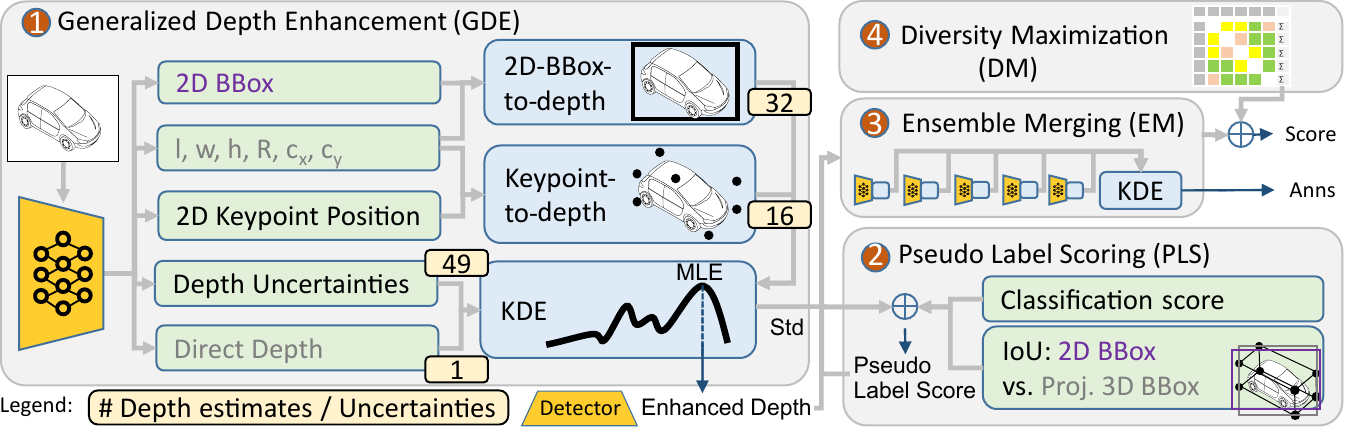}
  \caption{\textbf{Overview of our \ourmethod for pseudo label generation:}
  \textbf{(1) GDE}: We convert auxiliary 2D BBox and 2D corner keypoint predictions into depth estimates, which are then merged into a single depth estimate using a \ac{KDE}.
  \mbox{\textbf{(2) PLS}}: To identify the most accurate pseudo labels, we assess the standard deviation of the \ac{KDE} and evaluate 2D/3D BBox consistency.
  \textbf{(3) EM}: We enhance predictions further by employing an ensemble of five teacher models.
  \textbf{(4) DM}: We filter pseudo labels for quality and rotation diversity to optimize self-training.}
  \label{fig:method}
\end{minipage}
\hfill
\begin{minipage}[c]{0.28\linewidth}
\vspace{-0.5em}
\includegraphics[width=1.00\linewidth]{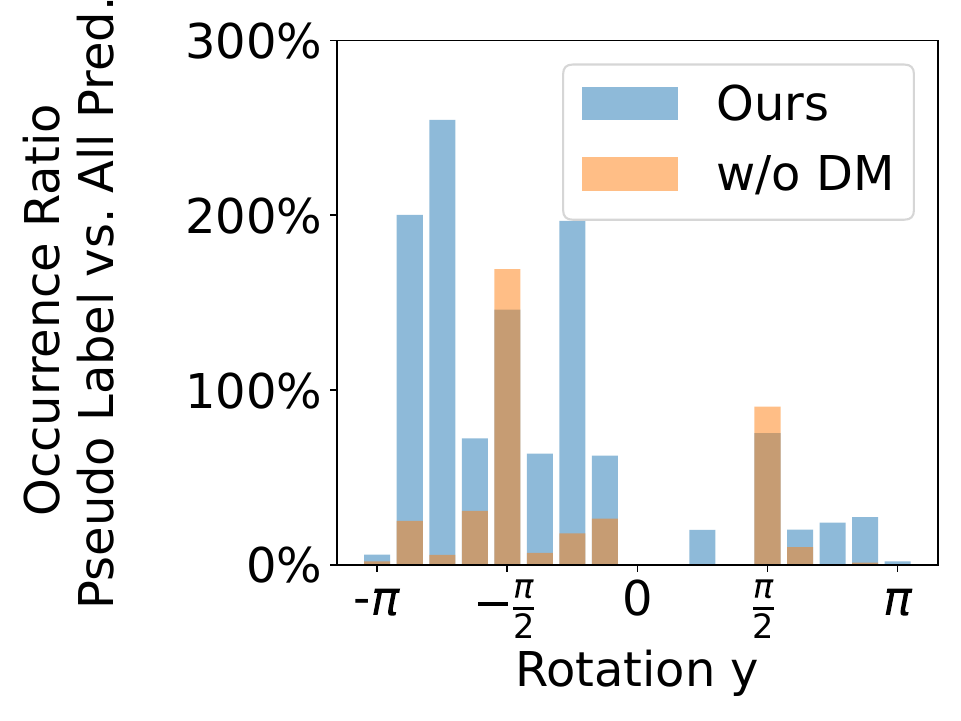}
  \caption{\textbf{Diversity Maximization (DM):} Without \ac{dm}, pseudo labels are concentrated at $\pm \frac{\pi}{2}$. With \ac{dm}, high-quality labels are more evenly distributed across all orientations (Lyft \cite{lyft} $\rightarrow$ KITTI \cite{kitti}).}
  \label{fig:rot_diversity}
\end{minipage}
\vspace{-0.5cm}
\end{figure*}

\subsection{Problem Statement}

The task of monocular 3D detection aims to predict 3D bounding boxes (BBoxes) ${b_1, \ldots, b_n}$ of objects from an RGB image $I$ with intrinsics $K \in \mathbb{R}^{3 \times 4}$. 
Each box $b_i$ is defined by its position $(x, y, z)$ relative to the camera, its dimensions $(l, w, h)$, its egocentric orientation matrix $R \in SO(3)$, and category $c \in \mathbb{N}$. 
For unsupervised domain adaptation, we use labeled source domain data $D_s = {\{I_j, K_j, \{b_{j,i}\}_{i=1}^{n_j}\}}_{j=1}^{M_s}$ and unlabeled target domain data $D_t = {\{I_j', K_j'\}}_{j=1}^{M_s}$ to maximize detection performance on the target test set.
Fig.~\ref{fig:method} shows our \ourmethod architecture and will be explained in the following.

\subsection{Baseline}
We use the state-of-the-art detector MonoCon \cite{monocon} with virtual depth \cite{omni3d} (denoted as MCVD) as our baseline.
MonoCon estimates offsets from the 2D object center to the projected 3D corner keypoints in pixel space as an auxiliary task. 
Our \ourmethod leverages these additional predictions in \cref{sec:generalized_depth_enhancement}.
Note that this auxiliary task can be easily integrated into other monocular 3D detectors.

\inparagraph{Generalization to new camera parameters}
To handle varying camera intrinsics across domains (\eg Lyft $\rightarrow$ KITTI), we use virtual depth $z_t$ \cite{omni3d} as training target instead of ground truth depth. 
The reason is that training MonoCon on the source domain only results in a score of 0 due to intrinsic shifts \cite{stmono3d}.
We normalizes depth with respect to camera intrinsics, with the added benefit of enabling scale augmentation: 
$z_t = s_{\text{aug}} \cdot \frac{f_{\text{ref}}}{f_{\text{img}}} \cdot z$, 
where $z$ is the ground truth depth, $\frac{f_{\text{ref}}}{f_{\text{img}}}$ is the ratio of the virtual to actual focal length, and $s_{\text{aug}}$ is the scale factor from augmentation. 
During inference, we unnormalize the depth by computing the inverse.
This approach enhances robustness to intrinsic changes and supports scale augmentation during training. The virtual focal length can be set arbitrarily. In our experiments, we set $f_{\text{ref}} = 700$.

\inparagraph{Support for $SO(3)$ rotations}
To effectively handle varying perspectives, our \ourmethod predicts $SO(3)$ rotations rather than a single rotation angle using an allocentric orientation representation  \cite{cdrone}. 
To maintain the properties of rotation matrices, we employ Gram-Schmidt orthogonalization. 
For further details, we refer to Brazil \etal \cite{omni3d}.

\inparagraph{Teacher-student training}
We leverage unlabeled target domain data through self-training in a teacher-student setup. We train an ensemble of five teacher models (same architecture, different seeds) on the source data and use the model ensemble to identify high-quality pseudo labels. Like Mix-Teaching \cite{mix-teaching}, we paste pseudo labels into empty images using border-cut, MixUp \cite{mixup}, and padding augmentation. To optimize resource efficiency, we perform only a single round of self-training. 

\subsection{Generalized Depth Enhancement (GDE)}
\label{sec:generalized_depth_enhancement}
Here, we introduce a novel method that converts auxiliary 2D predictions into 48 additional depth estimates and combines them into a single, unified estimate. Different from previous methods \cite{monoflex, monocd, monodde, gupnet}, which are limited to the egocentric car perspective, our method is designed to work with car, traffic, and drone views.

\inparagraph{Keypoint-to-depth}
Our baseline model predicts the 2D corner keypoint coordinates $(kp^{2D}_x, kp^{2D}_y)$ for each BBox as an auxiliary task (\textit{center to keypoint}). As they can also be obtained by projecting the 3D keypoint position $(kp^{3D}_x, kp^{3D}_y, kp^{3D}_z)$ into the image, we have: 
\begin{equation}
kp^{3D}_z \cdot [kp^{2D}_x, kp^{2D}_y, 1]^T = K [kp^{3D}_x, kp^{3D}_y, kp^{3D}_z, 1]^T. 
\label{eq:kp_proj}
\end{equation}
Solving \cref{eq:kp_proj} for depth yields two estimates for $kp^{3D}_z$ at each corner point, denoted as $z_x$ and $z_y$: 

\begin{equation}
z_x = \frac{f_x \cdot b_x + (c_x - kp^{2D}_x) \cdot b_z}{kp^{2D}_x - a_x}
\label{eq_zx}
\end{equation}

\begin{equation}
z_y = \frac{f_y \cdot b_y + (c_y - kp^{2D}_y) \cdot b_z}{kp^{2D}_y - a_y},
\label{eq_zy}
\end{equation}

\begin{equation}
\text{with } [b_x, b_y, b_z]^T = R \, ([r_x, r_y, r_z]^T \odot [l, w, h]^T)
\label{eq:b}
\end{equation}

where $(a_x, a_y)$ denotes the projected 3D object center, $R$ is the egocentric orientation matrix, $[l, w, h]$ is the object dimension, $(c_x, c_y)$ is the principal point, $f_x, f_y$ are the focal lengths, $\odot$ refers to the Hadamard product, and $(r_x, r_y, r_z)$ defines the keypoint location relative to the object with $r_x, r_y \in \{-0.5, 0.5\}$ and $r_z \in \{0, 1\}$. This formulation gives us 16 depth estimates and can be considered a generalization of MonoDDE's approach \cite{monodde} to $SO(3)$ rotation.

\inparagraph{2D-bounding-box-to-depth}
Estimating the 2D BBox is generally easier than estimating the 3D BBox.
Based on this insight,
we use the projected 3D BBox as the target for training the 2D BBox of our detector \cite{cdrone}. Assuming no truncation at the image boundaries, each side of the 2D bounding box must encompass at least one keypoint: one keypoint's x-coordinate will align with the left side and another with the right, while one keypoint's y-coordinate will coincide with the top and another with the bottom. Given that there are 8 corner points and 4 potential positions (left, right, bottom, top), there are 32 possible combinations, of which at least 4 are valid. We use this information to obtain depth estimates for each combination using \cref{eq_zx} and \cref{eq_zy}.

\inparagraph{Depth merging}
Only a subset of these $16+32=48$ auxiliary depth estimates will be accurate. \Eg consider the case, where we assume the keypoint is on the left side, but it is actually on the right side. In this case, the computed depth is negative. Instead of hand-designing a complex filtering mechanism, we let the network learn the aleatoric uncertainty that the estimated depth is correct. We add the following loss: 
\begin{equation}
    \mathcal{L}_{\text{aux}} = \frac{1}{48} \sum_{i=1}^{48} \sqrt{2} \cdot  \frac{|\hat{z}_i-z_{t}|}{\hat\sigma_i} + \log \hat\sigma_i
\end{equation}
where $\hat{z}_i$ and $\hat\sigma_i$ are the auxiliary depth estimates and implicitly learned uncertainties. Since $\hat z_i$ are obtained in closed-form from the network output, we only backpropagate through the uncertainties $\hat\sigma_i$.

The output depth is obtained by combining all depth estimates: the 48 estimations and an additional direct depth prediction (49 depth estimates in total). Previous methods have employed the weighted mean, which is equivalent to \ac{wbf} \cite{monoflex,monodde, weighted_box_fusion}. However, we empirically notice that it is not robust against outliers. Instead, we use kernel density estimation to merge the predictions. A \ac{KDE} can generate a continuous density function from sample points without requiring any additional parameters about the distribution of these points. For simplicity, we use a Gaussian kernel and infer the bandwidth parameter $h$ with Silverman's rule of thumb \cite{silverman}. We parametrize the points with our predicted depth values and set the weight to the exponential inverse uncertainty $w_i = \frac{\exp{\hat\sigma_i^{-1}}}{\sum_j^{49}\exp{\hat\sigma_j^{-1}}}$ : 
\begin{equation}
    \hat{f}_h(z) = \sum_{i=1}^{49} w_i \frac{1}{\sqrt{2\pi h^2}} \exp \frac{(z-z_i)^2}{2h^2}.
\end{equation}
We perform \ac{mle} on the KDE density to infer our merged depth because it is more robust to noise than alternatives like \ac{wbf}. This method ensures a high likelihood when multiple high-confidence depth predictions are closely grouped. In contrast, outliers, which generally have low confidence and few neighboring points, contribute to lower likelihood values.

\subsection{Pseudo Label Scoring (PLS)} \label{sec:pseudo_label_scoring}
Selecting only high-quality pseudo labels ensures the model learns reliable patterns from the target domain. While the default MonoCon \cite{monocon} score (class score times depth certainty) is adequate for evaluating overall prediction quality, it is not ideal for identifying the small subset of high-quality labels needed for self-training. We propose a new pseudo label score \mbox{$s_{\text{pseudo}} = (s_{\text{cls}} + s_{\text{KDE}} + s_{2D/3D})/3$}, where \(s_{\text{cls}}\) is the default class score of MonoCon \cite{monocon}, \mbox{\(s_{\text{KDE}} = \exp(-\text{STD}_{\text{KDE}}(z))\)} is the negative exponential of the KDE's standard deviation, and \(s_{2D/3D}\) is the \ac{iou} between the predicted 2D BBox and the projected 3D BBox. This new score emphasizes model consistency: when multiple high-confidence depth predictions are close together (high consistency), \(s_{\text{KDE}}\) is high, whereas when predictions are far apart (low consistency), \(s_{\text{KDE}}\) is low. 

\subsection{Ensemble Merging (EM)}
\label{sec:ensemble_merging}
Inspired by \cite{ms3d, ms3d++}, 
we train five models (same architecture, different seeds) and require that all models predict an object to qualify as a pseudo label. Unlike Mix-Teaching \cite{mix-teaching}, which uses the most confident model for pseudo annotations, we incorporate all predictions and merge them using a \ac{KDE} with \ac{mle}.
We use separate \ac{KDE}s for $x, y, z, w, h,$ and $l$ and employ $s_{pseudo}$ as the \ac{KDE} weight. As in previous work \cite{ms3d}, we take the rotation prediction from the most confident teacher.
We prefer a \ac{KDE} for  robustness to noise over methods like \ac{wbf} \cite{weighted_box_fusion}, as used in SOAP \cite{soap}.

\subsection{Diversity Maximization (DM)}
\label{sec:diversity_maximization}
With the above three stages, we can identify high-quality pseudo labels for self-training by enforcing a threshold \( s_{pseudo} > s_{min} \). However, we observe a tendency for specific configurations, such as cars driving forward, backward, left, or right relative to the camera, to be selected more frequently as pseudo labels. This pattern likely arises because these orientations are more prevalent in the dataset.
Unfortunately, this can result in oversampling and introduce a bias toward simpler cases.

To address this issue, we propose Diversity Maximization to enhance the rotation variety of pseudo labels. Let \( R_1, \ldots, R_n \) denote the allocentric \cite{omni3d} rotation matrices of the merged pseudo labels. We first measure the diversity of a pseudo label \(i\) with rotation matrix \(R_i\) \wrt the $N = 2,500$ most confident pseudo labels we would have chosen without diversity maximization. We can use the geodesic distance $d_{geo}$ as it measures the shortest path between two rotations on the \(SO(3)\) rotation manifold: 
\begin{equation}
d_{geo}(R_1, R_2) = \lVert \log (R_i^T R_j) \rVert.
\end{equation}
However, this approach encourages opposing directions, which lead to the maximum distance and are not desirable in our context. To address this, we employ the following recalibrated version with \(n_c=2\), which results in diversity peaks when being at a distance of \(45^\circ\) and \(135^\circ\) wrt. the average orientation:
\begin{equation}
\begin{aligned}
  d_{recalib}(R_1, R_2) &= \min\left(d , \frac{\pi}{n_c} - d\right) \text{ with} \\
  d &= \frac{d_{geo}(R_1, R_2)}{\sqrt{2}} \mod \frac{\pi}{n_c}.
\end{aligned}
\end{equation}

\noindent
Our diversity score is then defined as: 
\begin{equation}
s_{div} = \frac{1}{N-1} \sum_{j \neq i} d_{recalib}(R_i, R_j) \cdot \frac{n_c}{\pi}.
\end{equation}
We rank the pseudo labels using the formula \mbox{\((1 - w_{\text{div}}) \cdot \bar{s}_{\text{pseudo}} + w_{\text{div}} \cdot s_{\text{div}}\)}, where \(\bar{s}_{\text{pseudo}}\) is the average \(\bar{s}_{\text{pseudo}}\) score across all 5 teacher models. We then select the top $N$ objects as final pseudo labels for self-training. In our experiments, we set \(w_{\text{div}} = 0.2\).

\section{Experiments}

\inparagraph{Implementation Details} Our \ourmethod uses DLA-34 \cite{dla34} as the backbone, conducted on a single NVIDIA A40. 
We utilize the AdamW optimizer \cite{adamw} with a cosine learning rate scheduler, setting the learning rate to $2.25 \times 10^{-4}$ and weight decay to $10^{-5}$. 
We paste up to 8 pseudo labels per image. 

\subsection{Cross Dataset Domain Adaptation}
We evaluate our \ourmethod on the benchmarks~\cite{stmono3d,monogdg} for domain adaptation, which includes cross-dataset adaptation, synthetic-to-real transfer, and adaptation to various weather conditions. 
To ensure fairness, we exclude validation images from the unlabeled set, leaving 39,522 KITTI images.

\begin{table}
    \caption{\textbf{Domain adaptation results for dataset shifts (real).} 
    Results are shown for the car class. 
    MCVD$^*$ denotes MonoCon \cite{monocon} with Virtual Depth \cite{omni3d}.}
    \centering
    \begin{tabularx}{\linewidth}{
        @{}X
        *{3}{S[table-format=2.2]@{\hspace{0.5em}}}
        *{3}{S[table-format=2.2]@{\hspace{0.5em}}}}

    \toprule
    \multirow{2}{*}{Method} & \multicolumn{3}{c}{AP$^{50}_{BEV}$} & \multicolumn{3}{c}{AP$^{50}_{3D}$} \\
    \cmidrule(lr){2-4} \cmidrule(lr){5-7}
    & {Easy} & {Mod} & {Hard} & {Easy} & {Mod} & {Hard} \\ \midrule

    \multicolumn{7}{l}{\hspace{-9pt} \textbf{nuScenes \cite{nuscenes} $\rightarrow$ KITTI \cite{kitti}, R40}} \\
    Oracle & 64.86 & 49.17 & 44.35 & 60.81 & 45.72 & 39.96 \\ 
    \arrayrulecolor{black!20}\midrule
    STMono3D \cite{stmono3d} & 35.63 & 23.62 & 22.18 & 29.01 & 19.88 & 17.17 \\
    DGMono3D \cite{dgmono3d} & 34.22 & 28.99 & 27.82 & 28.77 & 24.82 & 23.67 \\
    MonoGDG \cite{monogdg} & 39.50 & 32.39 & 32.07 & 33.48 & 27.14 & 26.37 \\ 
    MCVD$^*$ {\scriptsize (Baseline)} & 32.96 & 23.98 & 20.02 & 21.31 & 14.89 & 17.36\\
    \ourmethod\ {\scriptsize (Ours)} & \textbf{52.39} & \textbf{36.24} & \textbf{34.95} & \textbf{42.80} & \textbf{32.24} & \textbf{27.36} \\ \arrayrulecolor{black}\midrule

    \multicolumn{7}{l}{\hspace{-9pt} \textbf{Lyft \cite{lyft} $\rightarrow$ KITTI \cite{kitti}, R11}} \\
    Oracle & 62.16 & 51.59 & 45.76 & 62.16 & 46.21 & 43.33 \\ \arrayrulecolor{black!20}\midrule
    STMono3D \cite{stmono3d} & 26.46 & 20.71 & 11.83 & 18.14 & 13.32 & 11.83 \\
    DGMono3D \cite{dgmono3d} & 36.18 & 28.30 & 22.23 & 30.03 & 23.28 & 22.23\\
    MonoGDG \cite{monogdg} & 38.47 & 30.89 & 29.58 & 32.48 & 26.02 & 24.96 \\ 
    MCVD$^*$ {\scriptsize (Baseline)} & 37.59 & 27.21 & 25.15 & 31.27 & 23.97 & 20.34 \\
    \ourmethod\ {\scriptsize (Ours)} & \textbf{59.54} & \textbf{43.70} & \textbf{38.07} & \textbf{50.54} & \textbf{36.33} & \textbf{30.50}\\ 
    \arrayrulecolor{black}\bottomrule
      
    \end{tabularx}
    \vspace{-0.2cm}
    \label{tab:exp_cross_dataset}
\end{table}
We train separately on NuScenes (204,894 images, Boston and Singapore) \cite{nuscenes} and Lyft (136,080 images, Palo Alto) \cite{lyft}, evaluating on KITTI val. (3,769 images, Germany) \cite{kitti}.  Both NuScenes and Lyft are subsampled to 1/4 of their data as in prior work \cite{stmono3d}. We report AP$_{BEV}$ and AP$_{3D}$ at an IoU of 0.5 for the car class (see \cref{tab:exp_cross_dataset}). We establish an oracle baseline by training MonoCon \cite{monocon} directly on the target dataset. Training exclusively on source data yields significantly degraded performance, highlighting the vulnerability of current state-of-the-art methods to cross-dataset domain shifts. Our approach effectively reduces this domain gap, retaining over 96\%, 85\%, and 83\% of the oracle performance on Lyft \cite{lyft} $\rightarrow$ KITTI \cite{kitti} AP$_{BEV}$. These results confirm that our method adapts 3D object detectors to the target domain more effectively and significantly outperforms existing methods. Notice that our method does not increase inference time, as it is only used to generate better labels for self-supervision.

\begin{table}
    \caption{\textbf{Domain generalization results for dataset shifts (synthetic + real):} nuScenes \cite{nuscenes} + PreSIL \cite{presil} $\rightarrow$ KITTI \cite{kitti}. 
    Results are shown for the car class (R40). 
    }
    \centering
    \begin{tabularx}{\linewidth}{
        @{}X
        *{3}{S[table-format=2.2]@{\hspace{0.5em}}}
        *{3}{S[table-format=2.2]@{\hspace{0.5em}}}}

    \toprule
    \multirow{2}{*}{Method} & \multicolumn{3}{c}{AP$^{50}_{BEV}$} & \multicolumn{3}{c}{AP$^{50}_{3D}$} \\
    \cmidrule(lr){2-4} \cmidrule(lr){5-7}
    & {Easy} & {Mod} & {Hard} & {Easy} & {Mod} & {Hard} \\ \midrule

    Oracle & 64.86 & 49.17 & 44.35 & 60.81 & 45.72 & 39.96 \\ \arrayrulecolor{black!20}\midrule
    
    STMono3D \cite{stmono3d} & 32.47 & 23.35 & 19.81 & 24.43 & 17.37 & 14.29\\
    DGMono3D \cite{dgmono3d} & 33.81 & 27.27 & 26.84 & 24.75 & 20.08 & 19.58 \\
    MonoGDG \cite{monogdg} & 38.35 & 29.48 & 28.71 & 31.55 & 23.31 & 22.25 \\ 
    MCVD$^*$ {\scriptsize (Baseline)} & 31.16 & 21.26 & 18.32 & 23.99 & 16.96 & 13.72 \\
    \ourmethod\ {\scriptsize (Ours)} & \textbf{56.45} & \textbf{35.62} & \textbf{28.83} & \textbf{49.44} & \textbf{30.33} & \textbf{25.12}\\ \arrayrulecolor{black}\bottomrule
      
    \end{tabularx}
    \vspace{-0.2cm}
    \label{tab:exp_synthetic}
\end{table}
\begin{table}
    \caption{\textbf{Domain adaptation results for dataset and weather shifts:} NuScenes \cite{nuscenes} + PreSIL \cite{presil} $\rightarrow$ KITTI fog/rain \cite{fog,rain}.
    Results are shown in AP$^{50}_{3D}$ R40 for cars.
    }
    \centering
    \begin{tabularx}{\linewidth}{
        @{}X
        *{3}{S[table-format=2.2]@{\hspace{0.5em}}}
        *{3}{S[table-format=2.2]@{\hspace{0.5em}}}}

    \toprule
    \multirow{2}{*}{Method} & \multicolumn{3}{c}{KITTI fog \cite{fog} AP$^{50}_{3D}$} & \multicolumn{3}{c}{KITTI rain \cite{rain}} \\
    \cmidrule(lr){2-4} \cmidrule(lr){5-7}
    & {Easy} & {Mod} & {Hard} & {Easy} & {Mod} & {Hard} \\ \midrule

    Oracle & 60.55 & 44.14 & 39.56 & 61.08 & 44.42 & 39.70 \\ \arrayrulecolor{black!30}\midrule
    STMono3D \cite{stmono3d} & 19.59 & 14.63 & 13.35 & 23.18 & 16.46 & 13.59 \\
    DGMono3D \cite{dgmono3d} & 17.64 & 12.52 & 11.84 & 22.30 & 15.92 & 15.26\\
    MonoGDG \cite{monogdg} & 23.58 & 15.97 & 15.28 & 27.69 & 18.33 & 17.79 \\ 
    MCVD$^{*}$ {\scriptsize (Baseline)} & 13.91 & 9.60 & 8.07 & 14.63 & 11.41 & 10.01\\
    \ourmethod\ {\scriptsize (Ours)} & \textbf{39.18} & \textbf{23.88} & \textbf{18.28} & \textbf{41.55} & \textbf{25.52} & \textbf{19.63}\\ \arrayrulecolor{black}\bottomrule
      
    \end{tabularx}

    \label{tab:exp_weather}
    \vspace{-0.2cm}
\end{table}

We then perform joint training on NuScenes \cite{nuscenes} and the synthetic dataset PreSIL ($51,075$ images, GTA V) \cite{presil}, and evaluate on KITTI val. \cite{kitti} (\cf \cref{tab:exp_synthetic}). 
Our \ourmethod performs consistently better than previous SOTA networks for the easy, moderate, and hard categories.

Additionally, we test generalization under different weather conditions. Training on NuScenes \cite{nuscenes} and PreSIL \cite{presil}, we evaluate on KITTI val. rain \cite{rain} and KITTI val fog \cite{fog} (\cf \cref{tab:exp_weather}). The baseline accuracy drops by 43.3\% and 32.7\%, whereas our approach only decreases by 21.3\% and 15.9\% for AP Mod. Notably, fog and rain images constitute less than 10\% of the unlabeled data here, as only the KITTI trainval set is provided with fog/rain \cite{fog,rain}. Despite this limitation, our \ourmethod significantly outperforms previous methods and the baseline, demonstrating its robustness and generalizability.

\begin{figure*}[t]
\begin{minipage}[c]{0.255\linewidth}
\hspace{-0.5em}
\centering
\begin{tabular}{c}
\includegraphics[height=66pt]{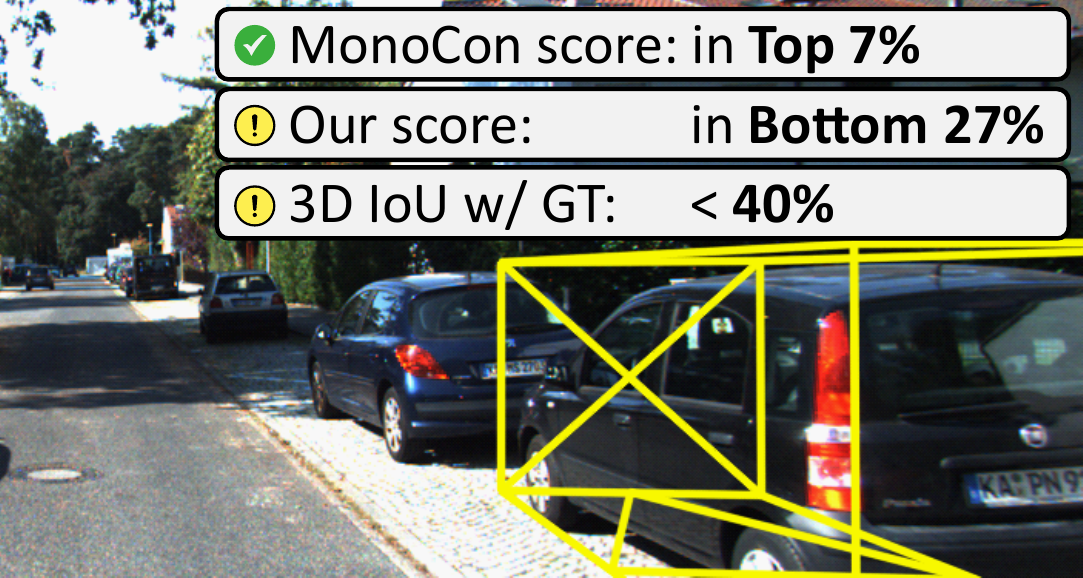}
\\
\includegraphics[height=66pt]{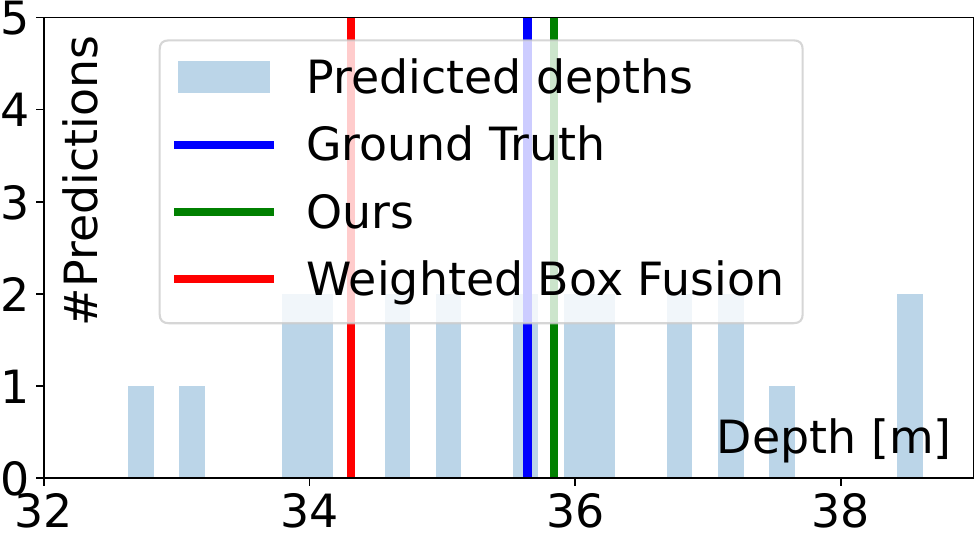}

\end{tabular}
\caption{\textbf{Top}: \ac{pls} filters inaccurate pseudo labels better than default scoring. \textbf{Bottom}: \ac{KDE}-based depth merging is more robust to outliers than \ac{wbf} \cite{weighted_box_fusion}.}
\label{fig:exp_ablation_score_generalized_depth_enhancement}

\label{fig:second_figure}
\end{minipage}
\hfill
\begin{minipage}[c]{0.73\linewidth}
\begin{tabular}{c c c}
\hspace{-1em}
\includegraphics[height=66pt]{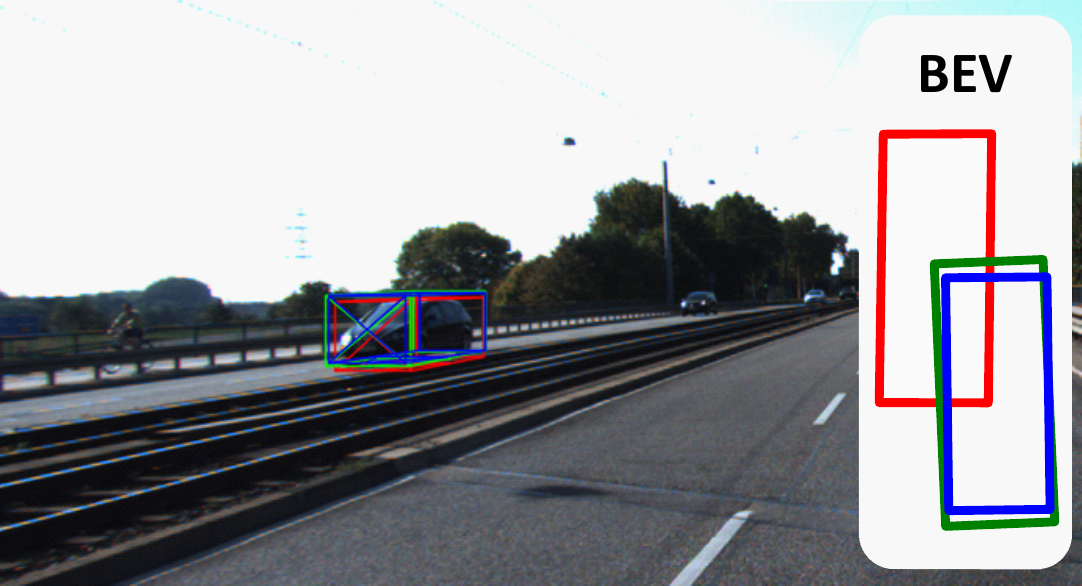}  & \hspace{-1em}
\includegraphics[height=66pt]{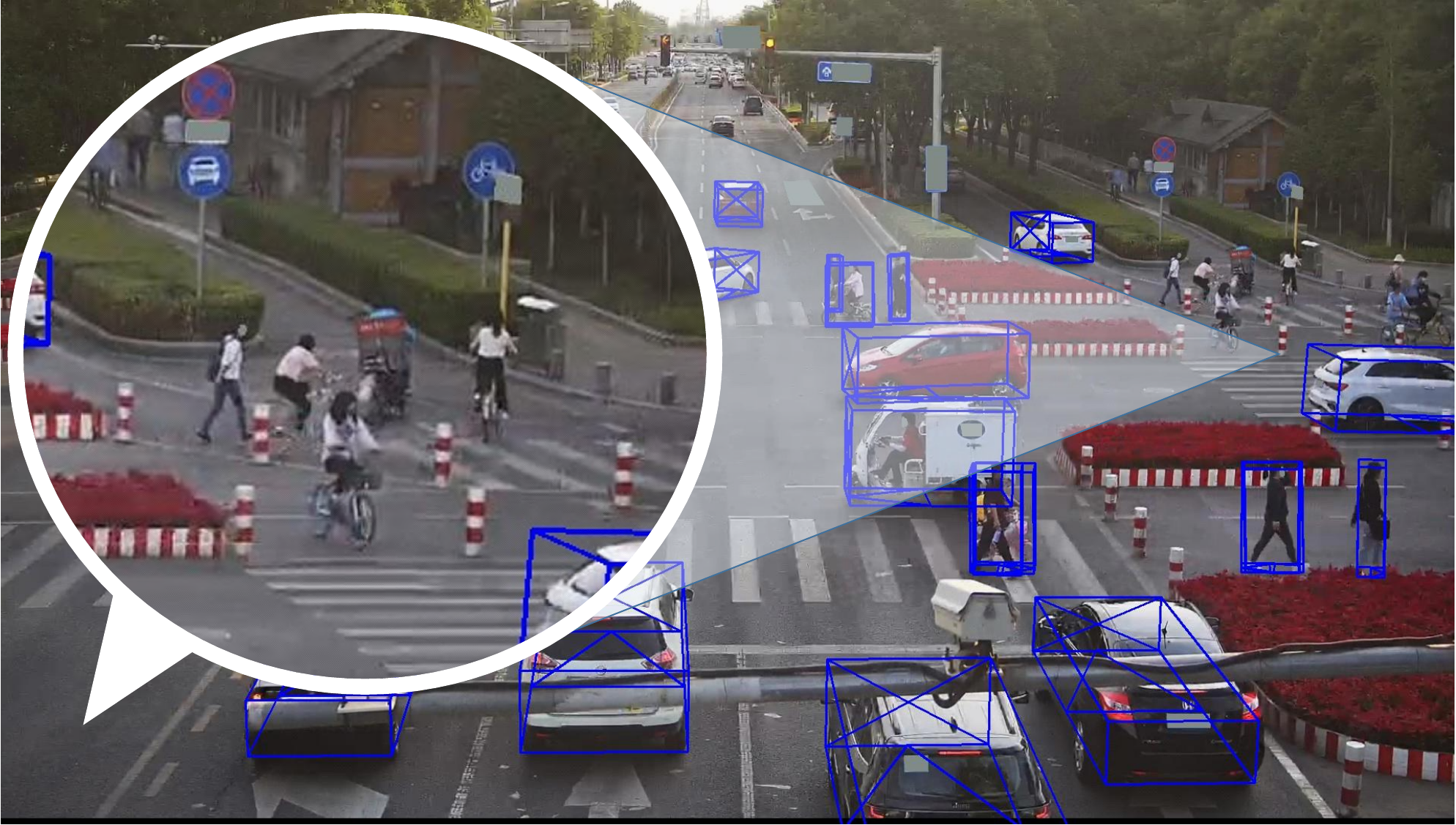} & \hspace{-1em}
\includegraphics[height=66pt]{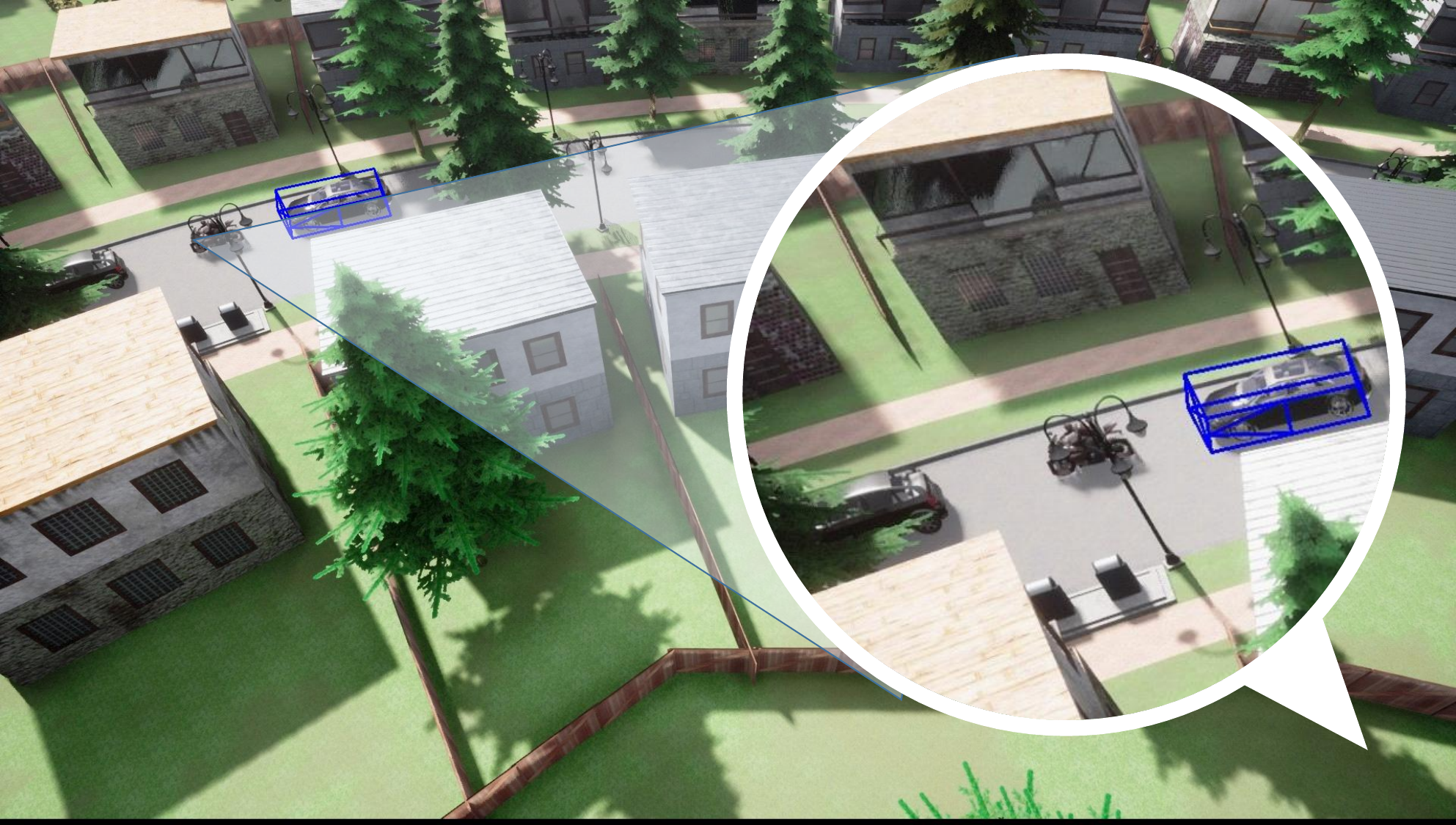}\\

\hspace{-1em}
\includegraphics[height=66pt]{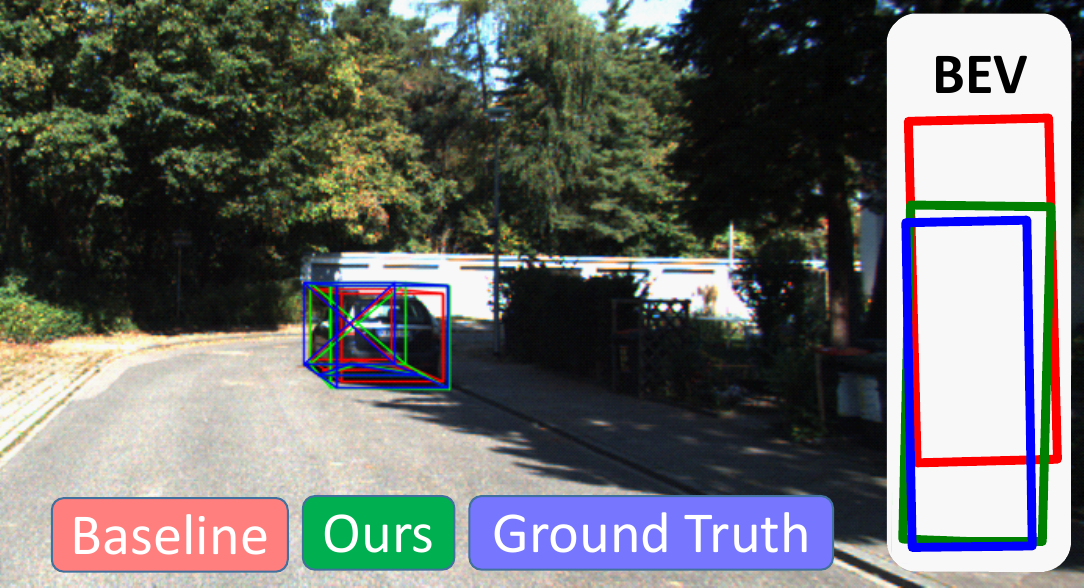} & \hspace{-1em}
\includegraphics[height=66pt]{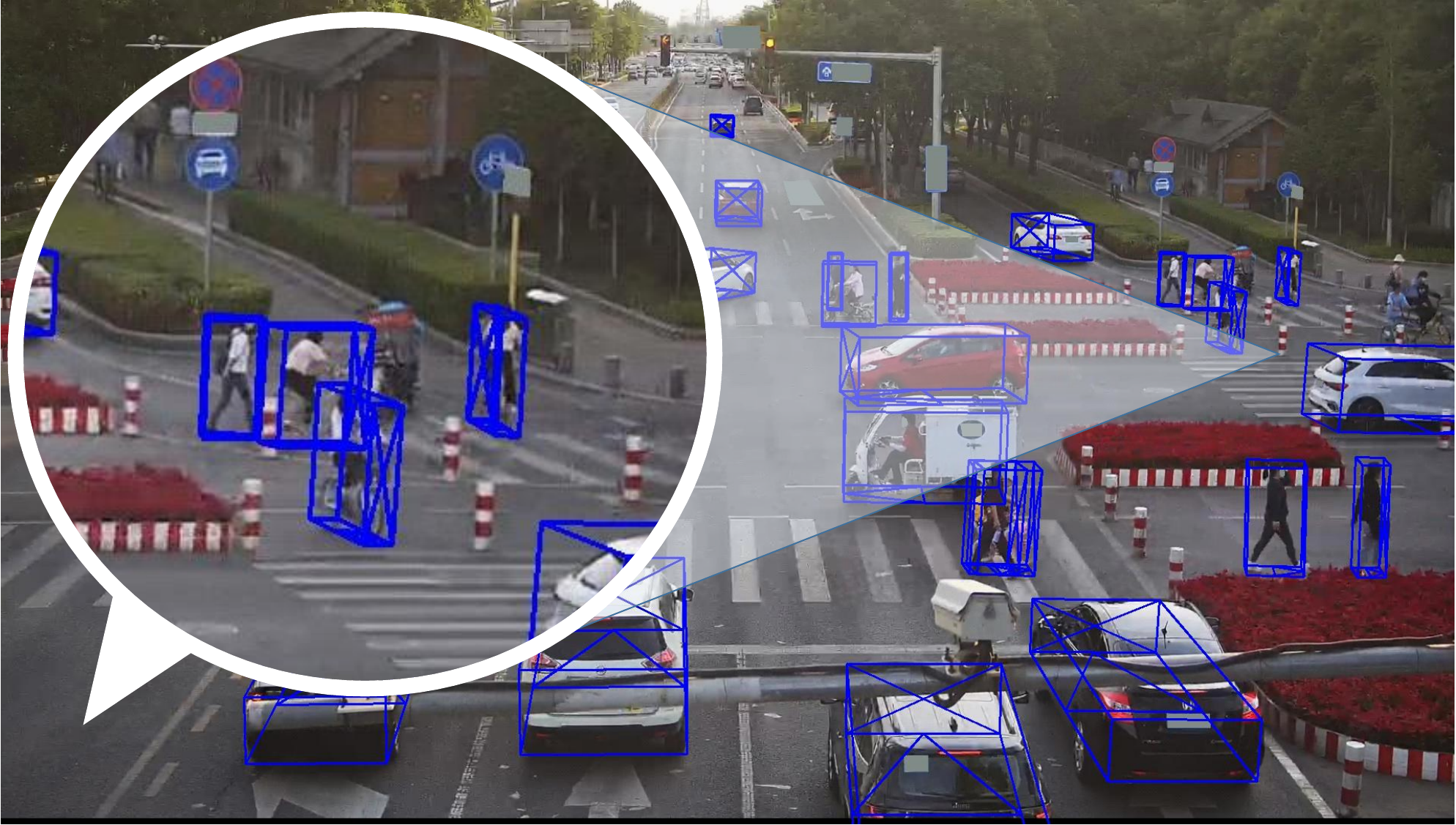} & \hspace{-1em}
\includegraphics[height=66pt]{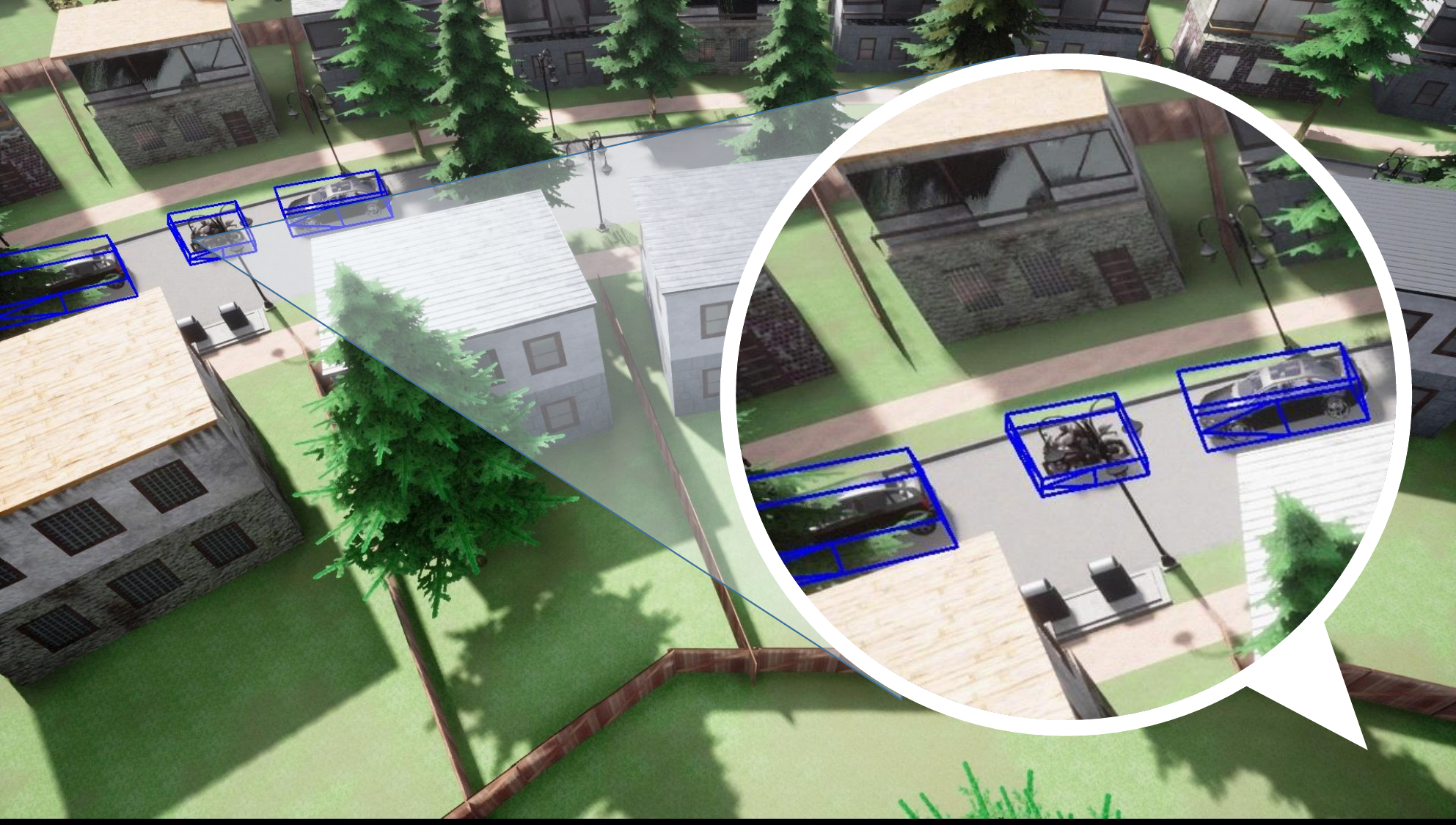}
\\
\end{tabular}
\caption{\textbf{Qualitative Results. } 
\textbf{Lyft \cite{lyft} $\rightarrow$ KITTI \cite{kitti}} (Column 1): Our pseudo label supervision allows \ourmethod to predict depth more accurately than MonoCon \cite{monocon}, as evident in the \ac{bev} visualization. \textbf{Rope3D \cite{rope3d} and CDrone \cite{cdrone}} (Columns 2 and 3): We compare our baseline MonoCon \cite{monocon} (row 1)
with \ourmethod (row 2). \ourmethod demonstrates superior detection of occluded and distant objects in new environments.}
\label{fig:main_qual_results}

\end{minipage}

\vspace{-0.5cm}
\end{figure*}

\subsection{Domain adaptation for traffic- and drone-view datasets}\label{sec:new_camera_locations}
\begin{table}
\caption{\ourmethod improves detection accuracy on the traffic- and drone-view datasets.
Results are reported for Car.}
    \centering
    \begin{tabularx}{1.0\linewidth}{
        @{}X 
        cr
        cr}

    \toprule
    Method & Rope3D \cite{rope3d} AP$_{3D}^{70}$ & CDrone \cite{cdrone} AP$_{3D}^{50}$ & \\ \midrule
    MCVD$^{*}$ {\scriptsize (Baseline)} & 4.73\,\,\,\,\, & 8.91\,\,\,\,\, \\
    \ourmethod\ {\scriptsize (Ours)} & \textbf{6.47}\,\,\,\,\, & \textbf{13.63}\,\,\,\,\, \\ 
    \rowcolor[HTML]{EFEFEF} Improvement & + 36.8\% & + 53.0\% \\ 
    \arrayrulecolor{black}\bottomrule
      
    \end{tabularx}
    \vspace{-0.5cm}
    \label{tab:exp_new_camera_location}
\end{table}

We evaluate \ourmethod on static camera datasets, Rope3D (traffic-view dataset with 45,008 images) \cite{rope3d}  and CDrone  (drone-view dataset with 37,800 images) \cite{cdrone}, both of which have cameras at a limited number of locations only. This makes it difficult to generalize to new camera locations when the training and test sets are from different locations \cite{cobev,mose}. To overcome this, we use sufficient unlabeled images from the test locations to boost accuracy.

Due to the limited images in raw test sets for Rope3D, we combine the raw training and validation set and sample half the locations for training. The remaining images are evenly divided between unlabeled data and evaluation sets, ensuring both contain the same locations. For CDrone, the training set serves as source data, and the combined validation and test set represents the target data, designating the first half of each video sequence as unlabeled. The last 25\% are used as test data, again ensuring both sets share the same locations. Results in \cref{tab:exp_new_camera_location} show a 36.8\% improvement on Rope3D and a 53.0\% improvement on CDrone, demonstrating the effectiveness of our \ourmethod 
when leveraging unlabeled data for diverse camera perspectives. Notably, pre-training on other datasets yielded no significant empirical benefits.


\subsection{Ablation Study}
\begin{table}
    \vspace{0.15cm}
    \caption{\textbf{Ablation Study}: Lyft \cite{lyft} $\rightarrow$ KITTI \cite{kitti}. We individually remove each component (\textbf{bold}) and sub-component from \ourmethod and report $AP_{3D}^{50}$ in R11 for Car.}
    \centering
    \begin{tabularx}{\linewidth}{
        @{}X
        *{3}{S[table-format=2.2]@{\hspace{0.5em}}}}

    \toprule
    Method & {Easy} & {Mod} & {Hard}\\ \midrule

    \ourmethod\ {\scriptsize (Ours)} & 50.54 & 36.33 & 30.49 \\
    \textbf{w/o Generalized Depth Enhancement (GDE)} & 36.17 & 26.40 & 23.91 \\
    w/o Keypoint to depth & 42.70 & 31.79 & 26.03 \\
    w/o 2D bbox to depth & 43.53 & 32.49 & 27.68 \\
    w/ Weighted Box Fusion (WBF) \cite{weighted_box_fusion,monodde} & 39.69 & 27.57 & 21.69 \\
    \textbf{w/o Pseudo Label Scoring (PLS)} & 41.02 & 30,73 & 26.82 \\
    w/o 2D/3D BBox consistency score & 47.27 & 34.61 & 29.37 \\
    w/ Mix-Teaching \cite{mix-teaching} ensemble score & 42.83 & 31.33 & 25.95\\
    \textbf{w/o Ensemble Merging (EM)} & 46.99 & 32.61 & 26.11 \\
    \textbf{w/o Diversity Maximization (DM)} & 43.19 & 28.50 & 25.68 \\
 \bottomrule
      
    \end{tabularx}
    \vspace{-0.15cm}
    \label{tab:exp_ablation}
\end{table}
We validate the effectiveness of our \ourmethod by systematically removing each component on the Lyft \cite{lyft} $\rightarrow$ KITTI \cite{kitti} benchmark (\cf \cref{tab:exp_ablation}). Among the primary components, \ac{gde}
proves to be the most influential. Removing its subcomponents \textit{keypoint-to-depth} or \textit{2D-bounding-box-to-depth} results in a 12.5\% and 10.6\% reduction in AP$_{3D}$ Mod, respectively, underscoring the necessity of incorporating both elements together. However, these additional depth estimates introduce significant outliers, which likely explains the 24.1\% performance drop in the Mod category when using a weighted average for fusion instead of a \ac{KDE} with \ac{mle} (\cf \cref{fig:exp_ablation_score_generalized_depth_enhancement} for a concrete example). 
Replacing \ac{pls} by the default score in MonoCon or with the ensemble score in Mix-Teaching reduces accuracy by 15.4\% and 13.8\% respectively (\cf example in \cref{fig:exp_ablation_score_generalized_depth_enhancement}), while removing the 2D/3D BBox consistency subscore reduces prediction quality by 4.7\%. 
To further validate our choice of pseudo label score, we compute the Spearman Rank Correlation between various scoring methods and the ground truth on the top 10\% of labels. Our pseudo label score achieves a correlation of 0.40, outperforming both the default MonoCon score (0.31) and the Mix-Teaching 3D \ac{iou} ensemble score (0.24).
Removing \ac{em} 
results in a 10.2\% decrease in accuracy, highlighting that combining multiple teacher predictions is more effective than relying solely on the most confident prediction, as in Mix-Teaching \cite{mix-teaching}.
Lastly, eliminating \ac{dm} 
results in a 21.6\% decrease in AP${3D}$ Mod, confirming the critical role of enhancing pseudo-label quality and diversity. The effect of this enhancement is further illustrated in \cref{fig:rot_diversity}, which demonstrates the increased variety of pseudo labels achieved with Diversity Maximization.

\subsection{Qualitative Results}
In \cref{fig:main_qual_results}, we present qualitative comparisons of our baseline MonoCon \cite{monocon} and \ourmethod. For Lyft \cite{lyft} $\rightarrow$ KITTI \cite{kitti}, \ourmethod demonstrates more accurate depth predictions. This improvement is expected, as \ourmethod performs self-training on a carefully selected subset of pseudo labels, which underwent both single-model depth enhancement through GDE and multi-model enhancement via \emph{EM}. Additionally, \ourmethod excels at detecting more objects in new camera locations for Rope3D \cite{rope3d} and CDrone \cite{cdrone} (see experimental setup in \cref{sec:new_camera_locations}).

\section{CONCLUSION}
We introduce \ourmethod, a carefully designed method to handle domain shift in monocular 3D object detection.
\ourmethod first converts auxiliary 2D predictions into depth estimates, significantly improving depth accuracy. 
Second, it enhances pseudo label scoring by incorporating model consistency measures.
Third, \ourmethod optimizes for both label quality and diversity to boost generalization. 
Experimental results demonstrate that \ourmethod significantly improves detection accuracy over existing methods. 
Additionally, \ourmethod shows superior generalization to new camera locations, underscoring its robustness and adaptability across diverse camera views.

\inparagraph{Limitations} 
Our evaluation is limited to outdoor scenes. While \ourmethod's components could theoretically apply to tasks like semi-supervised learning, few-shot learning, and general monocular 3D detection, we have not evaluated  these applications yet. We plan to explore this in future work.

\addtolength{\textheight}{0cm}    



\clearpage
\bibliographystyle{IEEEtran}
\bibliography{IEEEabrv,main}

\end{document}